\documentclass[acmtog]{acmart}
\acmSubmissionID{859}

\usepackage{booktabs} 

\usepackage[ruled]{algorithm2e} 

\SetAlFnt{\small}
\SetAlCapFnt{\small}
\SetAlCapNameFnt{\small}
\SetAlCapHSkip{0pt}

\copyrightyear{2024}
\acmYear{2024}
\setcopyright{rightsretained}
\acmConference[SIGGRAPH Conference Papers '24]{Special Interest Group on Computer Graphics and Interactive Techniques Conference Conference Papers '24}{July 27-August 1, 2024}{Denver, CO, USA}
\acmBooktitle{Special Interest Group on Computer Graphics and Interactive Techniques Conference Conference Papers '24 (SIGGRAPH Conference Papers '24), July 27-August 1, 2024, Denver, CO, USA}
\acmDOI{10.1145/3641519.3657478}
\acmISBN{979-8-4007-0525-0/24/07}

\begin{document}
\title{InvertAvatar: Incremental GAN Inversion for Generalized Head Avatars}

\author{XIAOCHEN ZHAO}
\authornote{Both authors contributed equally to the paper}
\orcid{0000-0001-8976-7723}
\email{zhaoxc19@mails.tsinghua.edu.cn}
\affiliation{%
  \institution{Tsinghua University}
  \city{Beijing}
  \country{China}
}

\author{JINGXIANG SUN}
\authornotemark[1]
\orcid{0000-0003-2966-9501}
\email{starkjxsun@gmail.com}
\affiliation{%
  \institution{Tsinghua University}
  \city{Beijing}
  \country{China}
}

\author{LIZHEN WANG}
\orcid{0000-0002-6674-9327}
\email{wang-lz@mail.tsinghua.edu.cn}
\affiliation{%
  \institution{Tsinghua University}
  \city{Beijing}
  \country{China}
}

\author{JINLI SUO}
\orcid{0000-0002-3426-1634}
\email{jlsuo@tsinghua.edu.cn}
\affiliation{%
  \institution{Tsinghua University}
  \city{Beijing}
  \country{China}
}

\author{YEBIN LIU}
\authornote{Corresponding authors}
\orcid{0000-0003-3215-0225}
\email{liuyebin@mail.tsinghua.edu.cn}
\affiliation{%
  \institution{Tsinghua University}
  \city{Beijing}
  \country{China}
}

\renewcommand{\shortauthors}{Zhao et al.}

\begin{abstract}
While high fidelity and efficiency are central to the creation of digital head avatars, recent methods relying on 2D or 3D generative models often experience limitations such as shape distortion, expression inaccuracy, and identity flickering. Additionally, existing one-shot inversion techniques fail to fully leverage multiple input images for detailed feature extraction. We propose a novel framework, \textbf{Incremental 3D GAN Inversion}, that enhances avatar reconstruction performance using an algorithm designed to increase the fidelity from multiple frames, resulting in improved reconstruction quality proportional to frame count. 
Our method introduces a unique animatable 3D GAN prior with two crucial modifications for enhanced expression controllability alongside an innovative neural texture encoder that categorizes texture feature spaces based on UV parameterization. Differentiating from traditional techniques, our architecture emphasizes pixel-aligned image-to-image translation, mitigating the need to learn correspondences between observation and canonical spaces. Furthermore, we incorporate ConvGRU-based recurrent networks for temporal data aggregation from multiple frames, boosting geometry and texture detail reconstruction. The proposed paradigm demonstrates state-of-the-art performance on one-shot and few-shot avatar animation tasks. Code will be available at https://github.com/XChenZ/invertAvatar.
\end{abstract}

%
%


%
%


\begin{teaserfigure}
\centering
  \includegraphics[width=\textwidth]{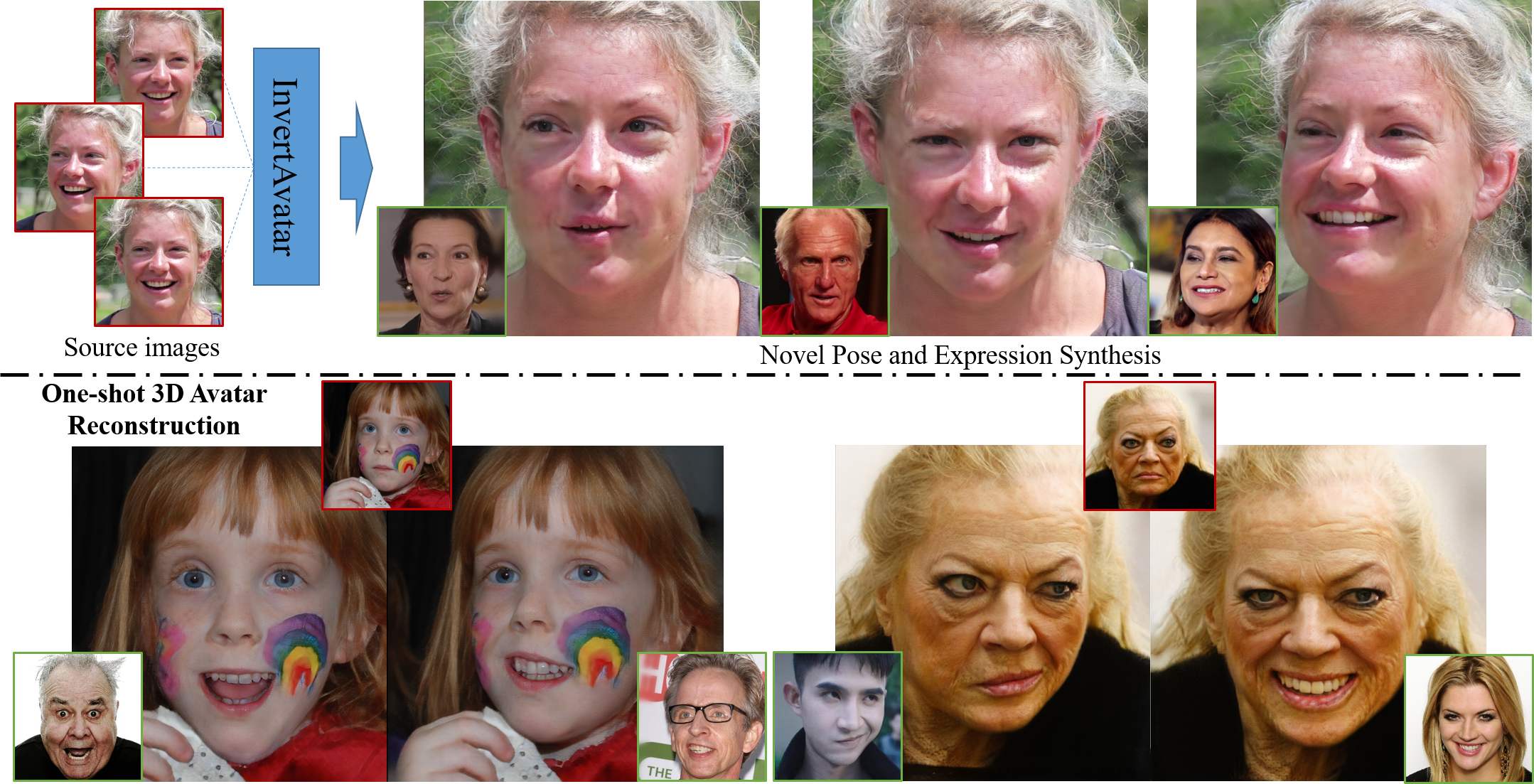}
  \caption{Given single or more source images, our method rapidly reconstructs photorealistic 3D facial avatars under one second, enabling precise control over full-head rotations and subtle expressions like lopsided grins and eye gazes.}
  \label{fig:teaser}
\end{teaserfigure}

\maketitle


\section{Introduction}
\label{sec:intro}

The creation of digital head avatars~\cite{gafni2021dynamic, zheng2022avatar, grassal2022neural, zhang2022fdnerf, zheng2023pointavatar, athar2022rignerf,deng2020disentangled, doukas2021headgan, ren2021pirenderer, wang2023styleavatar} has long been a task of interest in the fields of computer vision and graphics. The ability to interactively generate photo-realistic head avatars opens up new possibilities in areas such as augmented/virtual reality (AR/VR), 3D telepresence, and video conferencing. The primary aims when developing a digital head avatar are two-fold: high fidelity and efficiency. High fidelity involves preserving intricate details from input images, reconstructing occluded facial regions (especially in profile views), and accurately modeling dynamic features such as facial expressions, while maintaining consistent identity throughout animation. Efficiency demands a model that can adapt to varied unseen identities and motions without additional optimization at the inference stage. 

Recent literature showcases numerous attempts to create avatars from a single input image and perform animation with video frames. Several 2D generative models~\cite{deng2020disentangled, doukas2021headgan, ren2021pirenderer, wang2023styleavatar} perform image animation by incorporating the 3D Morphable Face Models (3DMM)~\cite{10.1145/311535.311556} into the portrait synthesis. These 2D-based methods achieve photo-realism but suffer from shape distortion during large motion due to a lack of geometry constraints. 

Towards better view consistency, many recent efforts~\cite{bai2023high,ma2023otavatar} leverage 3D GAN prior and decouple motion and appearance in latent space with video data. The main drawback of this paradigm is that the motion and appearance are naturally entangled in 3D GAN which leads to expression inaccuracy and identity flickering during animation. The other line of works~\cite{tang20233dfaceshop,sun2023next3d,li2023generalizable,wu2022anifacegan} directly learn a controllable 3D GAN from only 2D image collections, and perform GAN inversion to restore an avatar from one single image. The methods employing the controllable generative prior demonstrate efficiency, yet they are constrained by their dependency on a singular source image. A single image often falls short in fully representing the subject owing to factors such as occlusions and limited pose information. Utilizing diverse source images enriches appearance data, thereby diminishing the extent of hallucination required by an image generator. However, these one-shot inversion techniques do not directly enable the aggregation of partial observations across distinct frames to reconstruct more personalized details. Consequently, this represents a significant limitation to their capacity for detailed feature extraction.

In this paper, we address the challenge of enhancing avatar reconstruction performance when multiple input images are present.Our proposed algorithm is designed to elevate the fidelity of the recovered avatar both within and beyond the face region across multiple frames, with improvements in personal detail reconstruction scaling with the increase in frame count. 
To achieve this, we introduce a novel framework capable of efficient incremental 3D head avatar reconstruction, fusing multiple inputs without necessitating any additional optimization during inference from a sequence of monocular images. 

Specifically, we utilize an animatable 3D GAN prior, incorporating two key modifications to enhance expression controllability's efficiency and accuracy. Subsequently, we introduce an innovative neural texture encoder designed to categorize texture feature spaces in accordance with UV parameterization. This differentiates our approach from prevalent techniques that depend on networks to map an unposed image to a canonical tri-plane representation. Our proposed architecture emphasizes pixel-aligned image-to-image translation, which effectively eradicates the need for learning correspondences between observation and canonical spaces. This focus significantly improves its capacity for recovering fine details.

To accumulate temporal data from numerous frames, we further modify the one-shot inversion architecture to accommodate multiple inputs by leveraging ConvGRU-based~\cite{ballas2015delving} recurrent networks. This unique recurrent mechanism allows a flexible number of inputs and autonomously determines which information should be retained or discarded within streaming frames, thereby bolstering geometry and texture detail reconstruction capabilities.

To summarize, the contributions of our approach are:
\begin{itemize}
\item We provide a new paradigm for efficient avatar reconstruction for streaming data based on a concept called incremental 3D GAN inversion. The paradigm supports one to multiple source images inputs and achieves state-of-the-art performance on one- and few-shot facial avatar animation tasks.

\item We propose an innovative neural texture encoder alongside an animatable 3D GAN prior, allowing for the definition of a texture feature space on the UV parameterization which enhances the recovery of fine details by pixel-aligned nature.

\item We incorporate recurrent networks for temporal aggregation from multiple input frames extracted from monocular video, enhancing the quality of avatar reconstruction by considering sequential data.

\end{itemize}
\section{Related Works}
\label{sec:related_work}

\subsection{3D GAN inversion}

Building upon the achievements of 2D Generative Adversarial Network (GAN) inversion for image editing and manipulation~\cite{richardson2021encoding, tov2021e4e, dinh2021hyperinverter,alaluf2021restyle,wang2021HFGI}, current 3D GAN inversion methodologies~\cite{ko20233d3dganinversion,sun2022ide,lin20223dganinversion} perform a projection of specific images onto multiple instances within the pre-trained StyleGAN2 latent space~\cite{Karras2020stylegan2,abdal2019image2stylegan}. While the global latent space enhances 3D-aware portrait editing abilities, it results in a trade-off between reconstruction accuracy and editability, thereby complicating the accurate reconstruction of input images. Consequently, contemporary 3D GAN inversion techniques necessitate an estimated camera pose and slight generator weight adjustment~\cite{roich2021pivotal, Feng2022nearperfect, xie2023high} during testing to reconstruct out-of-domain input images accurately. In contrast to latent code inversion or tedious generator fine-tuning, recent efforts have shifted towards directly applying 3D GAN inversion on a triplane architecture for 3D GANs. For instance, TriPlaneNet~\cite{bhattarai2023triplanenet} presents a triplane encoder while Live3DPortrait~\cite{trevithick2023} integrates a Vision Transformer architecture into the triplane encoder, both showing substantial enhancements in robustness and 3D inversion quality for out-of-domain images. Furthermore, Control3Diff~\cite{gu2023learning} suggests a diffusion-based tri-plane encoder that demonstrates multimodal control capabilities. 

\subsection{2D Facial animation}

Leveraging the benefits of 3DMM, 2D facial animation tasks~\cite{gecer2019ganfit, thies2016face2face} are able to accurately and continuously model facial part deformations. However, these methods tend to lack fine-grained facial details, and grapple with representing non-facial areas that are beyond the scope of 3DMM, such as hair, teeth, eyes, and body.
To compensate for these limitations and generate more realistic facial details, subsequent research~\cite{kim2018deep, xu2020deep, thies2019deferred, gecer2018semi, fried2019text, tewari2020stylerig, deng2020disentangled, doukas2021headgan, ren2021pirenderer, wang2023styleavatar} has applied learned techniques atop 3DMM renderings. 
Getting rid of 3DMM, some methods~\cite{burkov2020neural, liang2022expressive, drobyshev2022megaportraits, wang2021one} are trained on the large-scale face video datasets containing rich identities and expressions, thus can be generalized to unseen motion and identity given just a single input image.
While these 2D-based approaches have successfully facilitated efficient and photorealistic animation, they do not model 3D geometry and hence fail to maintain stringent 3D consistency during significant head pose alterations.

\subsection{Neural 3D head avatars}

For strict 3D consistency, recent efforts~\cite{gafni2021dynamic, zheng2022avatar, grassal2022neural, zhang2022fdnerf, athar2022rignerf, xu2023avatarmav, zhao2023havatar, chen2023monogaussianavatar, xu2023gaussian} incorporate 3DMM into implicit representations~\cite{mildenhall2020nerf, sitzmann2019deepvoxels, kanazawa2018learning, peng2020convolutional, kerbl2023gaussian} to achieve view-consistent facial animation. Although these methods demonstrate realistic reconstruction results, they inefficiently learn separate networks for different identities and require thousands of frames from a specific individual for training. Another line of works~\cite{sun2023next3d, ma2023otavatar, yu2023nofa, li2023generalizable, bai2023high, li2023hide, xu2023omniavatar} focus on reconstructing 3D portraits in one-shot or few-shot manner incorporating 3D GAN prior or training in a generative manner. Next3D~\cite{sun2023next3d} learns a controllable 3D head avatar from random noise and reconstructs an avatar by PTI inversion. ~\cite{li2023generalizable} proposes to encode input images into a canonical branch for coarse geometry and texture, and introduces an additional appearance branch that captures personal details and an expression branch that modifies the reconstruction to the desired expression. These methods show efficient promising one-shot reconstruction but none of them are architected to accommodate the utilization of multiple source images as input. Contrastingly, our proposed framework is designed to leverage the complementary information inherent in multiple input images, thereby facilitating superior reconstruction quality.

\begin{figure}[t]
  \centering
  \includegraphics[width=0.45\textwidth]{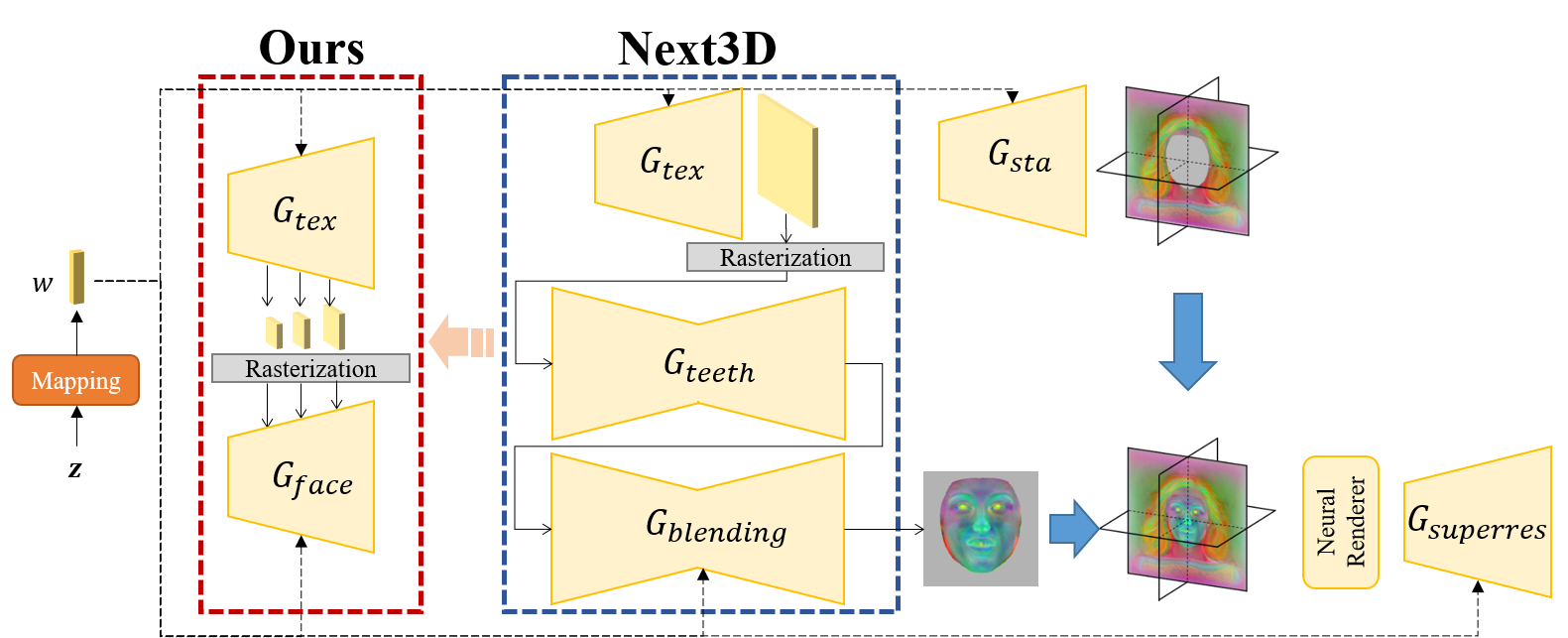}
  \caption{The architecture review of Next3D, along with the visualization of the modifications in our adopted 3D generative model.}
  \label{fig:next3d}
\end{figure}

\begin{figure*}[ht]
  \centering
  \includegraphics[width=\textwidth]{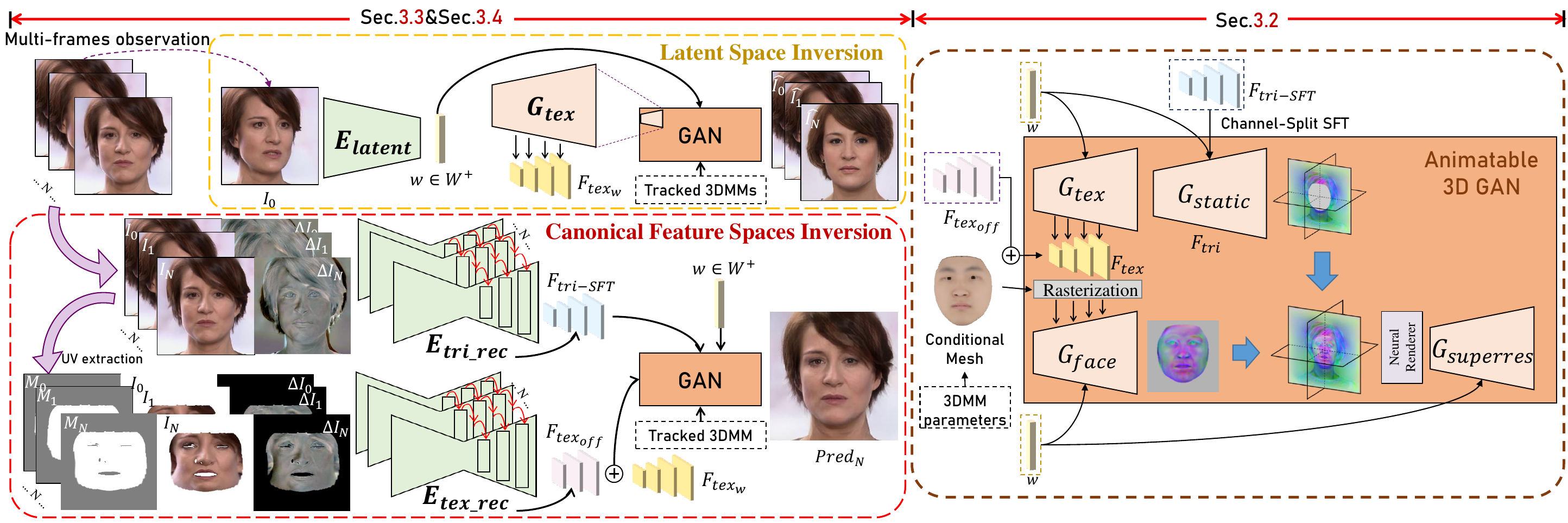}
  \caption{
   The left part illustrates our two-stage avatar reconstruction pipeline. The coarse stage "Latent Space Inversion" (Sec3.3) inverts the first frame in the GAN prior's $W+$ latent space with $E_{latent}$, forming an initial avatar. The fine stage performs offset prediction in canonical feature spaces (Sec3.3) and we specifically design recurrent networks $E_{tex\_rec}$ and $E_{tri\_rec}$ to aggregate temporal information, incrementally refining a high-fidelity avatar (Sec3.4). The architecture of our advanced animatable 3D generative model is depicted in the right box.
  }
  \label{fig:pipeline}
\end{figure*}
\section{Method}
\label{sec:method}

Our method innovatively harnesses the prior knowledge embedded in an animatable 3D generative model, integrating it with an incremental inversion architecture to enable rapid and high-fidelity 3D avatar reconstruction. Utilizing either a single source image or a short sequence of dynamic frames from a source video, our system is capable of reconstructing a 3D facial avatar, which is readily animatable through a 3D Morphable Face Model (3DMM)~\cite{wang2022faceverse}. In Sec.~\ref{subsec:next3d++}, we introduce our compact yet highly expressive animatable GAN, which provides a strong 3D prior. Sec.~\ref{subsec:oneshot} details our one-shot learning-based avatar reconstruction method, intricately designed to capitalize on the aforementioned GAN prior. Conclusively, in Section~\ref{subsec:gru}, we describe the employment of RNN-based temporal aggregation networks, which are instrumental in enhancing the shape and appearance precision of the reconstructed avatar by harnessing temporal data.

\subsection{Preliminary: Next3D}

Our chosen framework for animatable 3D GAN is Next3D~\cite{sun2023next3d}, which currently stands as the SOTA in generating high-quality, 3D consistent facial avatars from unstructured 2D images. The innovation within Next3D lies in its proposed 3D representation, termed \textit{Generative Texture-Rasterized Tri-planes}. Illustrated in Fig.~\ref{fig:next3d}, given a randomly sampled latent code $z$, the mapping network maps $z$ to an intermediate latent code $w$. This intermediate latent code modulates a StyleGAN generator to synthesize Generative Neural Textures atop parametric mesh templates. Subsequently, these are projected onto three orthogonally oriented feature planes via rasterization of a conditional facial 3DMM, thus establishing a tri-plane feature schema for volume rendering and 2D super-resolution ($G_{superres}$). This dual approach combines the nuanced expression control afforded by mesh-guided explicit deformation with the adaptive prowess of implicit volumetric representation. For more details please refer to \cite{sun2023next3d}.

\subsection{Compact and Expressive Animatable 3D GAN}
\label{subsec:next3d++}

Drawing from Next3D for our 3D facial avatar generator, we make the following improvements for our avatar reconstruction task. First, since 3DMM does not contain the inner mouth area, Next3D supplements and reintegrates the inner mouth features by two additional UNets, which leads to additional complexity and meory burden. To improve this, as shown in Fig.~\ref{fig:next3d}, we replace the two UNets with a single generator $G_{face}$ by conditioning $G_{face}$ with the multi-scale rasterized neural texture feature maps, which allows the inner mouth features to be gradually generated through multiple convolutional layers. Consequently, our more memory-efficient model design allows for larger neural rendering resolution of $128^2$.

Besides, we practically notice that the misalignment between conditional 3DMM and 2D image expressions can weaken their correlation during adversarial training, impeding the correct deformation of generated images. To rectify this, we employ Faceverse~\cite{wang2022faceverse}, a parametric model with more expression bases, to accurately fit subtle expression changes in 2D images. During training, we also supply the discriminator with the tracked landmarks map to bolster the consistency between synthesized images and the conditional mesh.

In the subsequent inversion and animation phases, we will initially fit Faceverse to the source portrait image or images of a person, capturing their expressions and head poses. Next, GAN inversion will be applied to reconstruct the corresponding head character. Finally, by changing the input expression parameters of FaceVerse, we gain the ability to animate the inverted avatar.

\subsection{One-shot Avatar Reconstruction}
\label{subsec:oneshot}

To extract maximum personal information from a single image with the reconstructed avatar, we introduce an effective one-shot 3D GAN inversion framework.
This framework utilizes a coarse-to-fine inversion architecture that carries out inversions in the latent and feature spaces of our animatable generative model separately. Due to the disentangled nature of the dynamic face and static head features, we divide the feature space into two specific canonical spaces: $F_{tex}$ defined by $G_{tex}$ (the canonical texture space), and $F_{tri}$ defined by $G_{static}$ (the canonical tri-plane space). The ultimate goal then becomes learning the mapping from the input portrait image to the facial avatar represented in these latent and canonical feature spaces.

In the first stage, we introduce a latent encoder $E_{latent}$ to project the image to the $W+$ latent space of the pretrained GAN. Specifically, given a source image $I$, we embed an inverted latent code $\hat{w}=E_{latent}(I)$ and initially generate a coarse facial avatar, represented by canonical features $F_{tex_w}=G_{tex}(\hat{w})$ and $F_{tri_w}=G_{static}(\hat{w})$. With the estimated pose and expression of the source image $I$, we can synthesize the fake image $\hat{I}=R(G_{face}(F_{tex_w}), F_{tri_w}  )$, where $R$ is the rendering block.

In the second stage, with the aim to offset the information loss instigated by $E_{latent}$, we formulate two separate image encoders $E_{tex}$ and $E_{tri}$. These are specifically designed to refine the feature maps in each of the two canonical spaces independently. Different from existing methods that rely on networks to map a posed image to the canonical tri-plane representation, benefiting from the neural texture architecture of our GAN prior, the face appearance represented by texture feature space is explicitly defined on the UV parameterization. Hence we design a U-Net network $E_{tex}$ to translate the observation to the canonical texture feature space in the UV domain, as shown in Fig.~\ref{fig:pipeline}\footnote{$E_{tex\_rec}$ in Fig.~\ref{fig:pipeline} refers to the encoder $E_{tex}$ equipped with a recurrent decoder. We describe it in Sec.~\ref{subsec:gru} in detail.}. Such a pixel-aligned image-to-image translation architecture helps eliminate the learning burden of correspondences between observation space and canonical space, making it concentrate on recovering fine details, as demonstrated in Sec.~\ref{subsec:abl_uvenc}. Specifically, the input of the $E_{tex}$ contains three maps, the source image $I$ and the residual map $\Delta{I} = I-\hat{I}$ that are projected to the UV plane, and a mask indicating the visibility region of the face. $E_{tex}$ outputs multi-resolution feature offsets $F_{tex_{off}}$ to compensate for the texture features $F_{tex_w}$ obtained in the first stage.

As for the static tri-plane features, we introduce an image-to-plane encoder $E_{tri}$ to predict multi-scale feature maps. Specifically, the multi-resolution features $F_{tri-SFT}$ output by $E_{tri}$ are used to spatially modulate $G_{static}$ with the Channel-Split SFT\footnote{Channel-Split Spatial Feature Transform is a layer tailored for the StyleGAN2 backbone to perform spatial modulation while preserving the generative prior.} layer~\cite{wang2021gfpgan} in a coarse-to-fine manner.

The training for all image encoders can also be divided into two stages. Firstly, we only train $E_{latent}$, with the loss terms following~\cite{tov2021e4e, richardson2021encoding}. Secondly, we simultaneously train $E_{tex}$ and $E_{tri}$. Referring to Live 3D Portrait~\cite{trevithick2023}, our training set includes the synthetic data and we supervise the training with many intermediate feature maps, as well as an adversarial loss to generate photo-realistic results. Please refer to our supplemental material for more details.

\subsection{Incremental GAN inversion}
\label{subsec:gru}

While the aforementioned GAN encoder effectively facilitates high-quality one-shot avatar reconstruction, a solitary image lacks sufficient pose and expression information to fully portray the shape and appearance. Therefore, we propose an approach termed \textit{Incremental GAN Inversion}: incrementally enhances the quality of GAN inversion while concurrently integrating temporal information.

We transition from a one-shot to a multi-frame inversion structure using temporal aggregation networks $E_{tex\_rec}$ and $E_{tri\_rec}$ equipped with recurrent decoders. Given an input video clip $\mathbf{I}^{N} = {\left \{ I_t \right \}}_0^N$, we aim at fusing the information of all frames to generate canonical representation $F_{tex}$ and $F_{tri}$. First, the first frame is input to the $E_{latent}$ for $W+$ space latent code $w$ and initial canonical features $F_{tex_w}$ and $F_{tri_w}$. According to the poses and expressions of the input sequence, we synthesize a fake image sequence and calculate the residual map for each frame $\Delta{I_t} = I_t-\hat{I_t}$. We sequentially input the observation and residual of each frame into the network $E_{tex\_rec}$ and $E_{tri\_rec}$ respectively on the UV domain and the picture domain.

Performing in a seq2one fashion, the recurrent decoders of $E_{tex\_rec}$ and $E_{tri\_tex}$ merge frame-wise feature maps in the temporal domain to predict one $F_{tri-SFT}$ and $F_{tex_{off}}$. 
Specifically, the recurrent decoder adopts ConvGRU~\cite{ballas2015delving} at each scale to aggregate temporal information, which is defined as:
\begin{equation}
\label{eq:gru}
\begin{split}
 &\  z_t, r_t = \sigma(Conv(f_t, h_{t-1})) \\
 &\  o_t = tanh(Conv(f_t, r_t*h_{t-1})) \\
 &\  h_t = z_t * h_{t-1} + (1-z_t)*o_t
\end{split}
\end{equation}
where operator $*$ denotes element-wise product. For each ConvGRU block, the feature map of one frame $f_t$ is input to calculate the update weights map $z_t$ and reset weights map $r_t$, to merge the current frame’s information into hidden state $h_t$. With the initial recurrent state $h_0$ all zero tensor, after inputting all frames in turn, the final fully updated hidden state serves as the output.

Contrary to systems employing straightforward feed forwarding of multiple frames as auxiliary inputs, our recurrent mechanism accommodates a flexible number of inputs. It intelligently learns the information to keep or discard from streaming data, thereby circumventing the constraints of fixed-size input windows and interval-based updates.

\begin{figure}[t]
  \centering
  \includegraphics[width=0.45\textwidth]{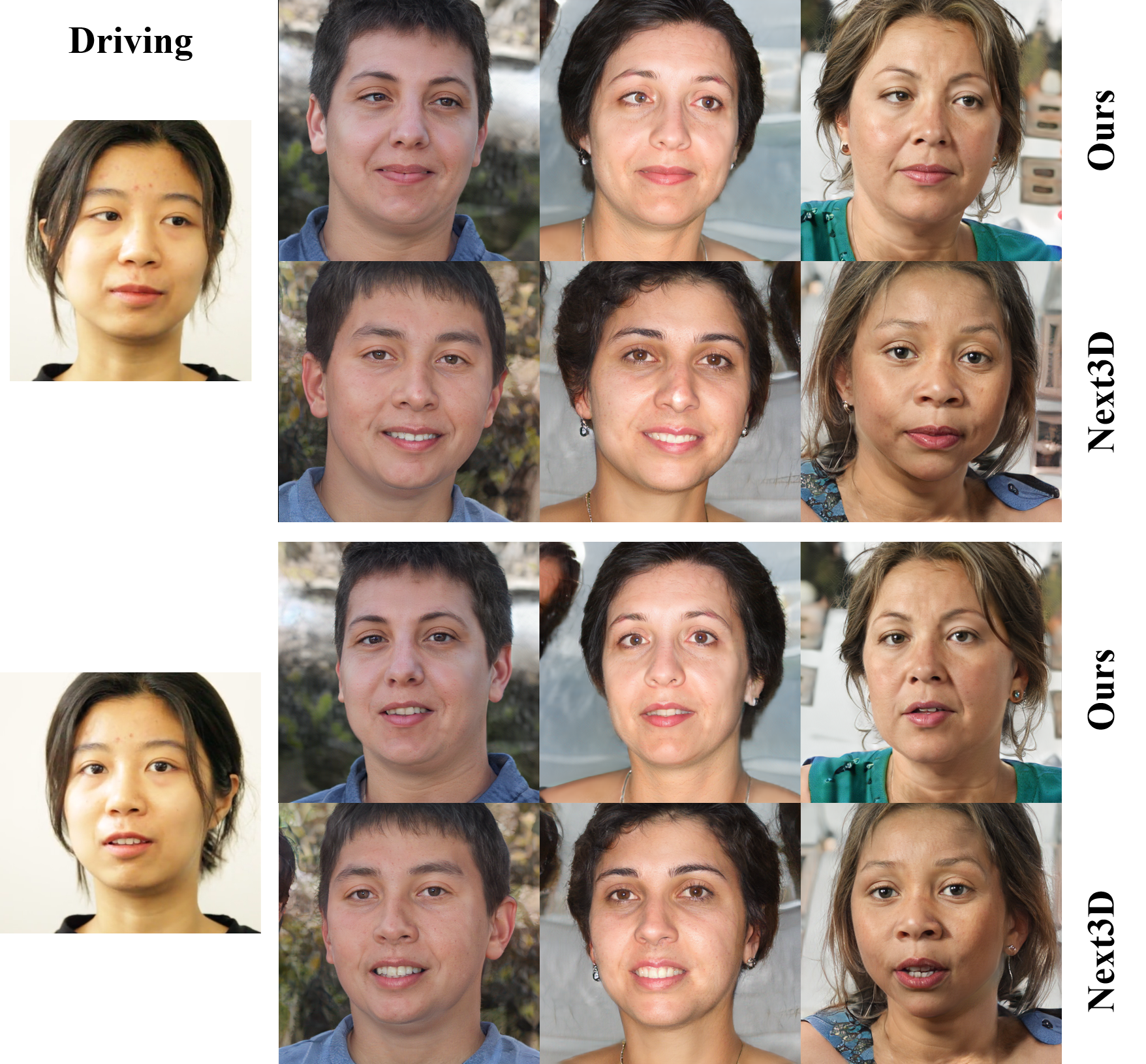}
  \caption{
Compare our animatable 3D generative model with Next3D~\cite{sun2023next3d}. We extract frames from a driving video clip and use the estimated facial model parameters to animate sampled random virtual avatars. Please zoom in and also refer to our video for more clear comparisons.
}
  \label{fig:cmp_next3d}
\end{figure}

\section{Experiments}
\label{sec:experiment}

\subsection{Experiment Setting}

\begin{table}[t]
    \centering
    \caption{Quantitative comparison with Next3D on image synthesis quality, animation accuracy, identity preservation and memory consumption.}
    \begin{tabular}{c|cccc}
    \hline
        ~ & FID$\downarrow$ & AKD$\downarrow$ & CSIM$\uparrow$ & Params \\ \hline
        Ours & 4.1  & 0.057 & 0.853 & 84.21M \\
        Next3D  & 3.9  & 0.248 & 0.841 & 164.81M \\ \hline
    \end{tabular}

\label{table:cmp_next3d}
\end{table}

\begin{table*}[t]
\caption{
Quantitative Evaluation for one-shot avatar reconstruction and animation.
}
    \centering
    \begin{tabular}{cccccccccccc}
    \hline
        ~ & \multicolumn{5}{c}{Same-ID on HDTF} & ~ & \multicolumn{2}{c}{Cross-ID on HDTF} & ~ & \multicolumn{2}{c}{Cross-ID on CelebA}\\ \cline{2-6} \cline{8-9} \cline{11-12}
        Method & PSNR $\uparrow$ & L1 $\downarrow$ & LPIPS $\downarrow$ & CSIM $\uparrow$ & AKD $\downarrow$ & ~ & CSIM $\uparrow$ & AKD $\downarrow$ & ~ & CSIM $\uparrow$ & AKD $\downarrow$ \\ \hline
        Ours & \textbf{23.27} & \textbf{0.029} & \textbf{0.111} & \textbf{0.873} & \textbf{0.018} & ~ & \textbf{0.759} & \textbf{0.041} & ~ & \textbf{0.755} & \textbf{0.069}  \\ 
        OTAvatar & 22.85 & 0.035 & 0.133 & 0.794 & 0.136 & ~ & 0.669 & 0.420 & ~ & 0.569 & 2.050  \\ 
        Next3D-PTI & 22.05 & 0.038 & 0.128 & 0.850& 0.312 & ~ & 0.651 & 0.577 & ~ & 0.648 & 0.788   \\ 
        StyleHEAT & 22.32& 0.036& 0.130& 0.827& 0.109& ~ & 0.710& 0.216& ~ & 0.604& 0.348\\ 
        ROME & 22.91& 0.034& 0.127& 0.832& 0.120& ~ & 0.722& 0.239& ~ & 0.682& 0.299\\ \hline
    \end{tabular}

\label{table:comp_os}
\end{table*}

\paragraph{Baseline methods.}
Firstly, we compare the utilized animatable 3D GAN prior with Next3D~\cite{sun2023next3d}. Then, we mainly compare our method with image-driven-based generalized head avatar reenactment methods, including a 2D face reenactment method~\cite{yin2022styleheat} and two 3D avatar synthesis methods~\cite{ma2023otavatar, Khakhulin2022ROME}. We also compare with an optimization-based baseline Next3D-PTI, which incorporates an animatable 3D GAN~\cite{sun2023next3d} and pivotal tuning inversion~\cite{roich2021pivotal}. We do not make comparisons with person-specific head avatar methods, since these multi-shot methods cannot generalize to different identities.

\paragraph{Evaluation Metrics.}
We evaluate generation, reconstruction and reenactment quality using Frechet Inception Distance (FID)~\cite{heusel2017fid}, peak signal-to-noise ratio (PSNR), Learned Perceptual Image Patch Similarity (LPIPS)~\cite{zhang2018perceptual} for synthetic quality, cosine similarity (CSIM)~\cite{deng2018arcface} for identity preservation, and average keypoint distance (AKD) using a SOTA off-the-shelf facial landmarks detector\footnote{https://github.com/google/mediapipe} for reenactment quality.

\begin{table*}[h]
\caption{
Quantitative Evaluation on VFHQ-Test using multiple source images.
}
    \centering
    \begin{tabular}{cccccccccccc}
    \hline
        ~ & \multicolumn{5}{c}{2 source frames} & ~ & \multicolumn{5}{c}{4 source frames} \\ \cline{2-6} \cline{8-12} 
        ~ & PSNR $\uparrow$ & LPIPS $\downarrow$ & AKD $\downarrow$ & CSIM $\uparrow$ & Time(s) & ~  & PSNR $\uparrow$ & LPIPS $\downarrow$ & AKD $\downarrow$ & CSIM $\uparrow$ & Time(s) \\ \hline
        Ours & \textbf{23.39} & \textbf{0.170} & \textbf{0.024} & \textbf{0.867}  & 0.20 & ~  & \textbf{23.58} & \textbf{0.162} & \textbf{0.022} &  \textbf{0.873} & 0.32   \\ 
        OTAvatar & 17.39 & 0.281 & 0.696 & 0.736 & 50 & ~  & 17.76 & 0.276 & 0.614 & 0.721 & 100  \\ 
        Ours-OS & 22.40 & 0.202 & 0.029 & 0.846 & 0.36 & ~ & 22.78 & 0.202 & 0.026 & 0.839 & 0.71  \\ 
        Next3D-PTI & 20.23 & 0.199 &0.637 & 0.821 & 150 & ~  & 20.74 & 0.1771 & 0.287 & 0.845 & 150  \\ \hline
    \end{tabular}
\label{table:comp_fs}
\end{table*}

\subsection{Comparisons}

\subsubsection{Animatable 3D GAN prior}
To validate the superiority of our compact and expressive animatable 3D GAN, we compare our generative model with Next3D. 
Fig.~\ref{fig:cmp_next3d} and Tab.~\ref{table:cmp_next3d} demonstrate qualitative and quantitative results. 
We measure image quality with FID between the entire FFHQ dataset and 50k generated images using randomly sampled latent codes, camera poses and estimated facial model parameters. Then we randomly sample 100 identities and animate each with randomly sampled 100 images from FFHQ, synthesizing 10000 images. AKD is calculated between the driving images and the animated generated ones, and CSIM is calculated for each identity.
The comparison proves that, with roughly half the number of parameters, our generative model achieves the similar synthesis quality as Next3D and significantly improves animation accuracy. In the \textbf{supplemental video}, the video animation results present that Next3D suffers from shaking facial appearance and texture sticking artifacts, while our method achieves better temporal- and view-consistency.

\subsubsection{One-shot Avatar Reconstruction}
We evaluate same-identity and cross-identity reenactment using various methods. For self-reenactment, the first frame of HDTF test videos served as the source, with the next 500 frames as the driving video. In cross-identity cases, the first 200 frames of one HDTF clip compose the driving video, and the first frame of other videos as source images. Besides, following~\cite{yin2022styleheat}, We also transfer motions from videos to CelebA-HQ monocular images~\cite{karras2018progressive}, using test videos from HDTF and VFHQ, creating 100,000 synthesized images from 1000 CelebA-HQ images and 25 video clips~\footnote{In addition to 19 videos from the testing split of HDTF, we also select 6 videos with rich expression changes from VFHQ as supplements.}. Without ground truth for cross-reenactment, we relied on CSIM and AKD metrics for evaluation.

\begin{figure*}[ht]
  \centering
  \includegraphics[width=\textwidth]{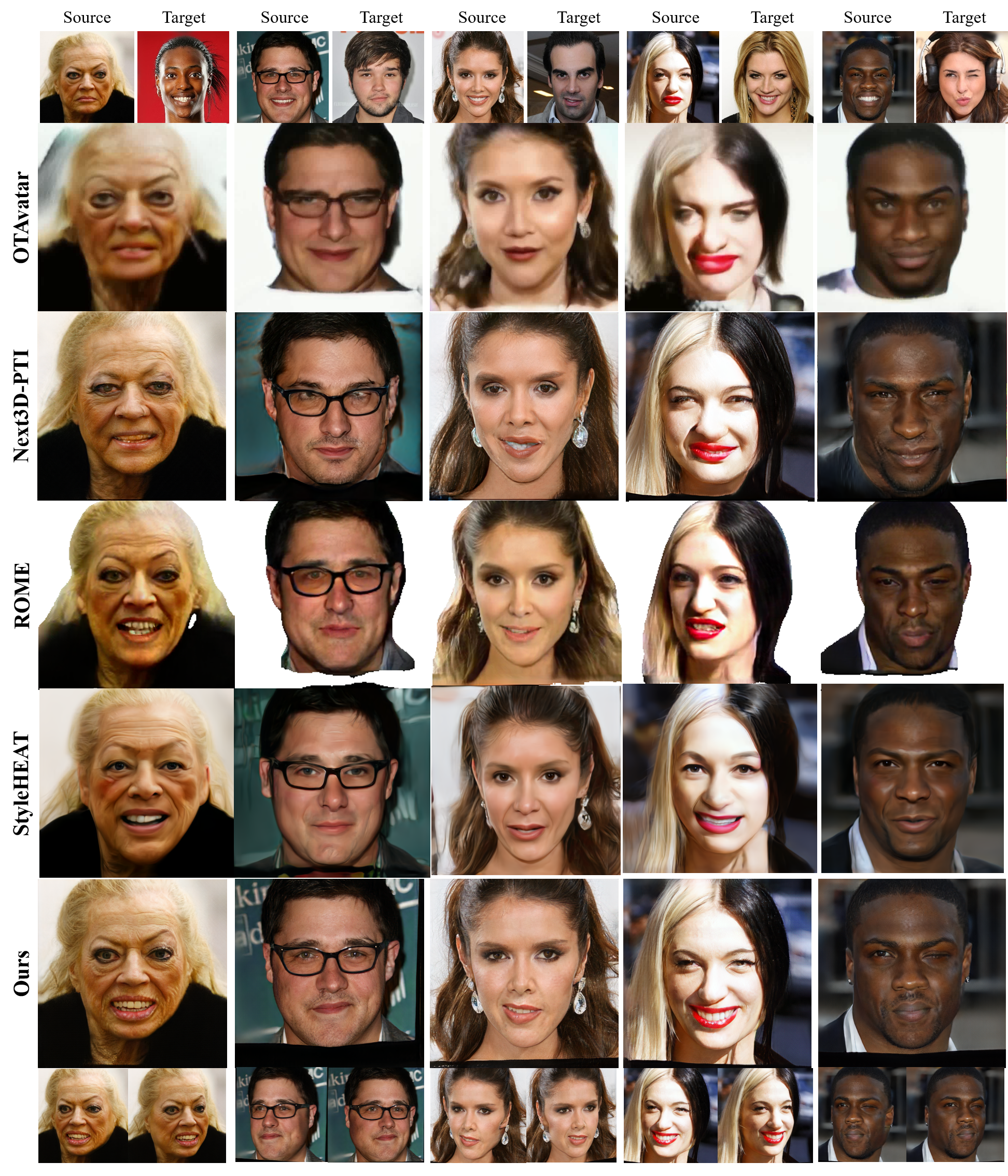}
  \caption{Qualitative comparison on the CelebA-HQ for one-shot avatar reconstruction. The results demonstrate the superior performance of our method in terms of realistic appearance recovery and fine-grained expression control.
}
  \label{fig:one-shot}
\end{figure*}

Tab.~\ref{table:comp_os} presents quantitative evaluations on the HDTF and CelebA-HQ datasets, while Fig.~\ref{fig:one-shot} showcases the qualitative results of cross-identity reenactment on the CelebA-HQ dataset. Our method shows superior generalization on these unseen datasets, surpassing others in identity preservation, appearance recovery, and expression control. It outperforms baselines in capturing fine details such as hair strands and earrings. Additionally, it also excels in expression transfer, offering precise control over nuanced expressions like asymmetrical mouth movements and eye gazes.
In comparison, ROME~\cite{Khakhulin2022ROME} fails to capture high-frequency details, StyleHeat~\cite{yin2022styleheat} struggles with inner mouth synthesis, and OTAvatar~\cite{ma2023otavatar} is unable to generate photo-realistic facial avatars with compact latent vector encoding. Although Next3D-PTI can reconstruct avatars with test-time optimization, it falls short in precise expression control.

\subsubsection{Few-shot Avatar Reconstruction}

Without readily available benchmarks for few-shot avatar reconstruction, we adapt one-shot methods, both our method and OTAvatar~\cite{ma2023otavatar}, to accommodate multiple source images. This adaptation involves reconstructing canonical features from each source image, followed by averaging these features to derive a canonical space representation. Specifically, for OTAvatar, we compute the mean identity latent code. For our one-shot method, we averaged the feature maps $F_{tex}$ and $F_{tri}$ in both two canonical spaces.

Due to the diverse expressions and poses in VFHQ~\cite{xie2022vfhq}, we evaluate using a test dataset of 100 VFHQ videos, each with 300 frames, excluding training set subjects. We sample input frames evenly from the initial 200 frames and use the last 100 frames for performance evaluation.

\begin{figure*}[ht]
  \centering
  \includegraphics[width=\textwidth]{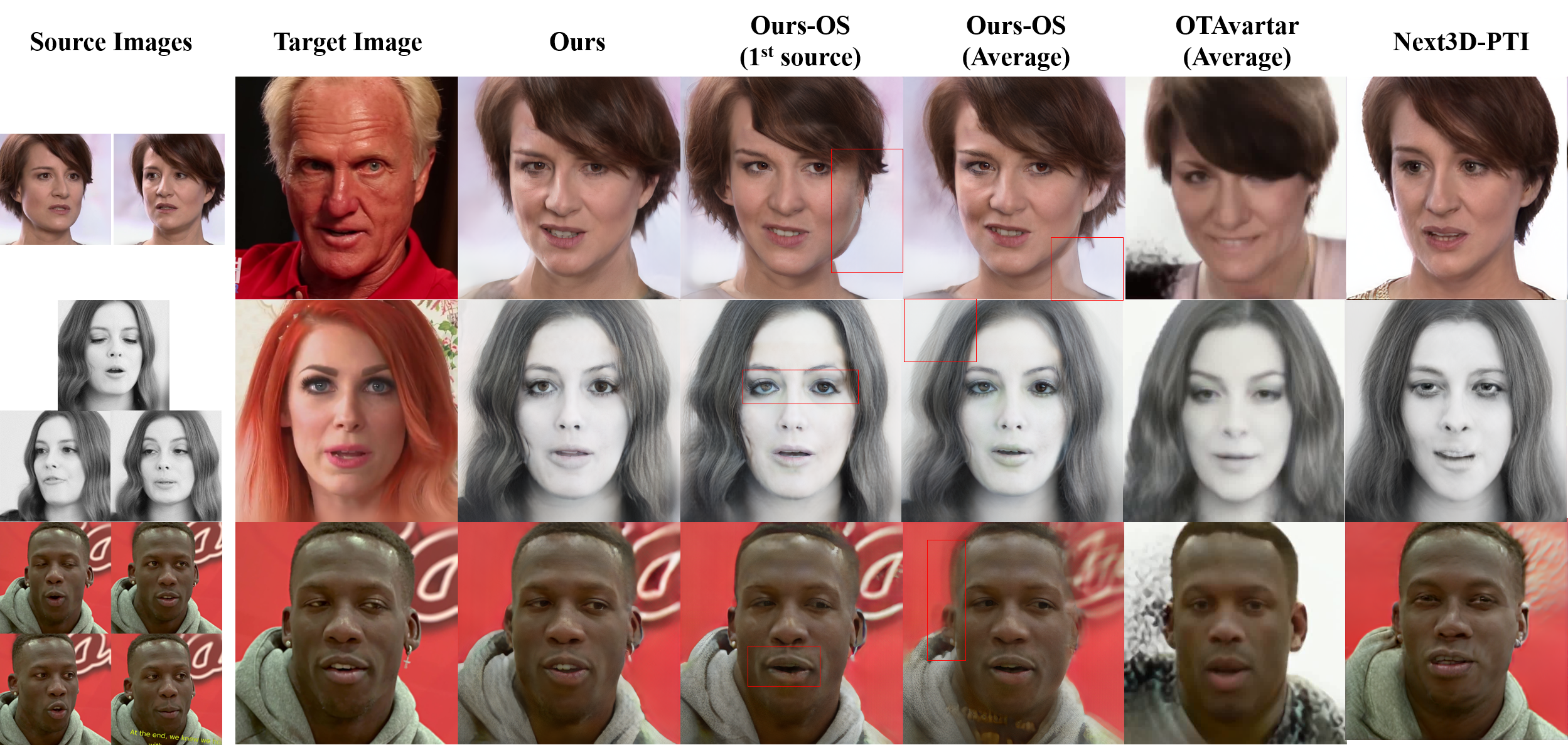}
  \caption{Qualitative comparison on the VFHQ-Test for few-shot avatar reconstruction and animation. Our method integrates multi-frame observations to improve shape reconstruction and texture recovery, as well as retaining high fidelity. 
  Though Ours-OS can also achieve photo-realism, it only uses the first image, and it may sometimes learn an inaccurate appearance for unobserved areas.}
  \label{fig:few-shot}
\end{figure*}

Tab.~\ref{table:comp_fs} and Fig.~\ref{fig:few-shot} display the results obtained with varying numbers of input frames. Our incremental GAN inversion framework, in comparison to the one-shot inversion using only the first frame, adeptly integrates multi-frame data to fill in information gaps present in a single image, thus enhancing the geometry and texture details. For example, this is evident in the improved depiction of hair shape (row 1, Fig.~\ref{fig:few-shot}), pupil color (row 2), and teeth refinement (row 3). Adapted one-shot methods, despite warping features from each source image to the canonical space, suffer from the misalignment of these inverted features, leading to blurry and inaccurate outputs. Our approach mitigates these issues, thanks to our RNN-based temporal aggregation network's effectiveness in merging sequential features, which is critical for achieving realistic detail recovery, as seen in the hair strands in Fig.~\ref{fig:teaser}.

For efficiency, OTAvatar\cite{ma2023otavatar} and Next3D-PTI require lengthy optimization for novel identities. In contrast, our learning-based solution achieves rapid avatar reconstruction within one second, with the time cost increasing only marginally with the number of input frames.

\subsection{Ablation Studies}
\label{subsec:ablation}

\subsubsection{Recurrent mechanism for temporal aggregation}
\label{subsec:abl_rnn}

To evaluate our recurrent mechanism, we establish a fixed-window baseline termed ``ConvFusion'', which replaces each GRU block with a convolutional block to compress temporal features. Due to its fixed input capacity, ``ConvFusion'' averages the inverted canonical features when handling longer sequences.

\begin{figure}[t]
  \centering
  \includegraphics[width=0.5\textwidth]{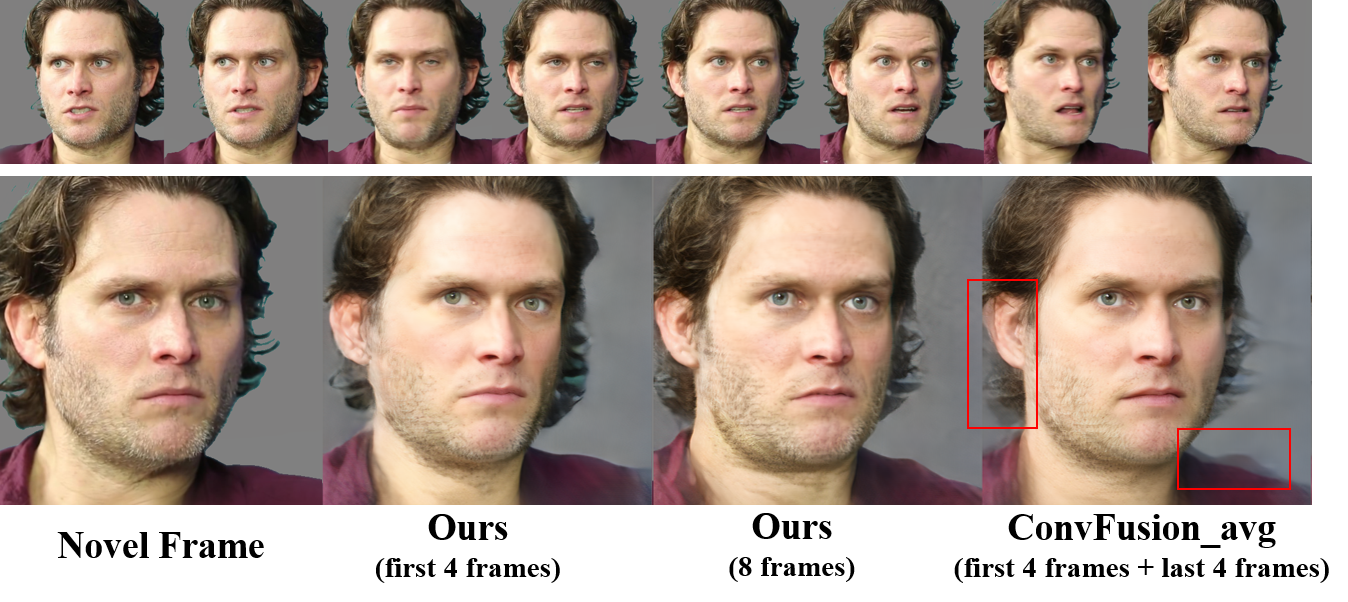}
  \caption{
  With a continuous input video stream, our method incrementally refines facial shape and texture details, in contrast to the fixed-window baseline ``ConvFusion\_avg'', which tends to produce blurry outcomes.
}
  \label{fig:abl_8frames}
\end{figure}

Fig.~\ref{fig:abl_8frames} demonstrates the use of an 8-frame video clip to mimic an online stream. Our method shows incremental improvements in the reconstructed shape and appearance with the increase in frames. In contrast, ``ConvFusion'' lacks the ability to discern which information to retain or discard, and a simple averaging strategy results in a blurred feature space and inferior rendering results.

\begin{figure}[t]
  \centering
  \includegraphics[width=0.4\textwidth]{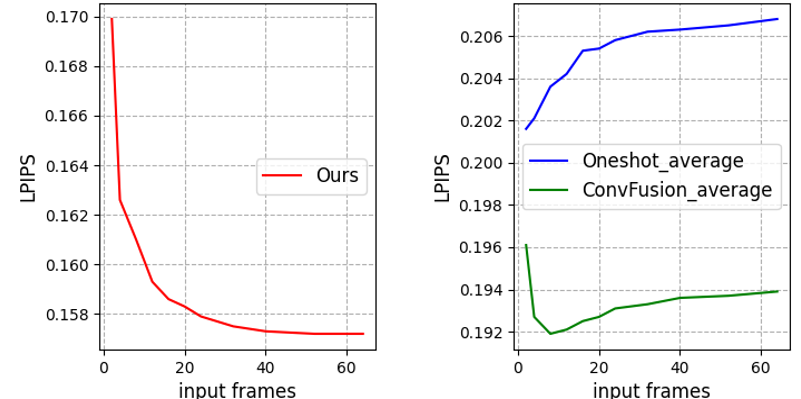}
  \caption{LPIPS over the number of input frames on VFHQ-test. With longer sequences, our method shows improving and converging metrics, affirming its proficiency in long-term temporal aggregation, unlike fixed-window baselines that degrade as source image count consistently increases.
}
  \label{fig:abl_multiT}
\end{figure}

A similar pattern emerges in Fig.~\ref{fig:abl_multiT}, where we simulate processing offline video sequences, assessing performance against the number of input frames.
On the VFHQ-test, with denser sampling in the first 200-frame sequence, The error of ``ConvFusion'' initially decreases but then increases after the frame count exceeds its fixed-window size. Conversely, our recurrent framework's performance improves incrementally with additional frames and then remains stable.

\subsubsection{Encoder design for canonical space inversion }
\label{subsec:abl_uvenc}

\begin{figure}[t]
  \centering
  \includegraphics[width=0.5\textwidth]{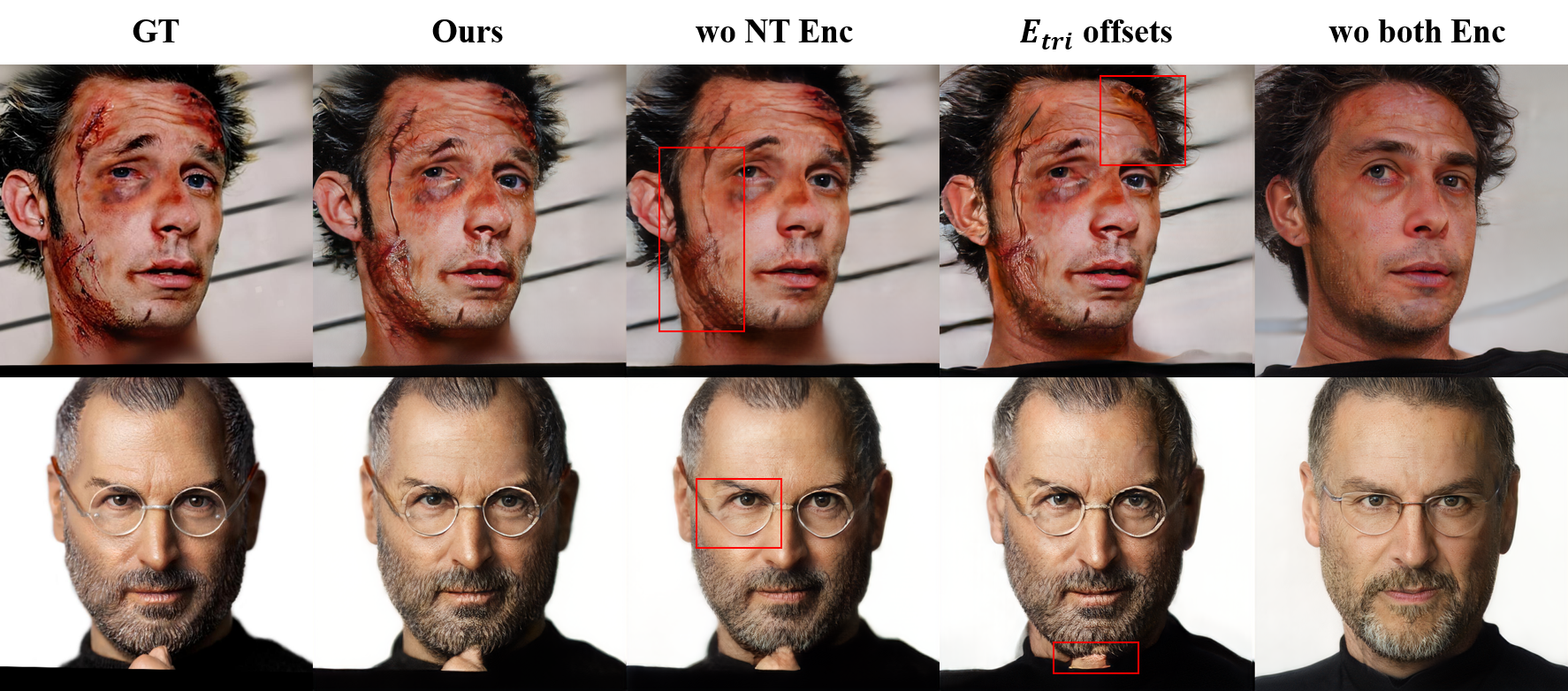}
  \caption{Qualitative ablation study results on the encoder design for canonical space inversion. Zoom in for the best view.
}
  \label{fig:abl_enc}
\end{figure}

\paragraph{Remove the second stage inversion} 
We set a baseline only utilizing $E_{latent}$. As shown in Fig.~\ref{fig:abl_enc} and Tab.~\ref{table:abl_enc}, without both $E_{tex}$ and $E_{tri}$, the reconstructed portrait cannot recover the identity.

\paragraph{Prediction of tri-plane SFT parameters.}
We introduce a baseline approach that utilizes $E_{tri}$ to directly predict feature offsets.
As illustrated in Fig.~\ref{fig:abl_enc} and Tab.~\ref{table:abl_enc}, our approach of predicting SFT parameters for modulating $G_{sta}$ lessens the impact of the initial coarse facial avatar, thereby reducing artifacts and improving fidelity.  

\paragraph{Neural Texture Encoder.} 
We set a baseline with posed images as inputs to $E_{tex}$. As shown in Fig.~\ref{fig:abl_enc} and Tab.~\ref{table:abl_enc}, our neural texture encoder, designed to prioritize pixel-aligned image-to-image translation on the UV domain, is vital for faithful facial detail recovery.

\begin{table}[t]
\caption{Ablation studies on CelebA-HQ for portrait reconstruction, to validate the effectiveness of our encoder design. We sample 1000 images to calculate metrics.
}
    \centering
    \begin{tabular}{c|cccc}
    \hline
        ~ & LPIPS$\downarrow$ & PSNR$\uparrow$ & CSIM$\uparrow$ & FID$\downarrow$ \\ \hline
        Ours  & \textbf{0.180}  & \textbf{25.34} & \textbf{0.933} & \textbf{23.25} \\
        wo both Enc  & 0.393  & 17.67 & 0.809 & 26.42 \\
        wo NT Enc  & 0.201  & 24.35 & 0.910 & 28.23 \\
        $E_{tri}$ offsets & 0.244 & 21.97 & 0.889 & 34.36 \\ \hline
    \end{tabular}

\label{table:abl_enc}
\end{table}

\section{Discussion and Conclusion}

\noindent\textbf{Limitation.} Dependent on a parametric model to control facial expressions, our method is error-prone to tracking errors and struggles with extreme expressions beyond the model's capability, like frowning and sticking tongue out.
Additionally, synthesizing stable, clear dynamic lower teeth remains challenging. Future advancements in more robust facial models capable of tracking these aspects are likely to enhance the performance.

\noindent\textbf{Potential Social Impact.}
Considering our method can reconstruct a vivid personalized head character using a single image in less than one second, it may be used for generating ``deepfakes'', which should be addressed carefully before deploying the technology.

\noindent\textbf{Conclusion.} We present \textit{Incremental 3D GAN Inversion}, a feedforward pipeline for generalized 3D head avatar creation, supporting one or multiple image inputs. We employ an animatable 3D GAN prior and a novel UV-aligned neural texture encoder for detail recovery and ConvGRU-based recurrent networks for effective temporal fusing of streaming data. Experiments have demonstrated our method outperforms the state-of-the-art approaches in both one-shot and few-shot avatar animation tasks. We believe our method with a flexible number of input frames will make progress for the GAN inversion methods.

\begin{acks}
This paper is supported by National Key R$\&$D Program of China (2022YFF0902200), the NSFC project No.62125107.
\end{acks}

\bibliographystyle{ACM-Reference-Format}
\bibliography{sample-acmtog-SIGGRAPH-submission}

\appendix
\clearpage
\setcounter{page}{1}

\section{Implementation Details}
In this section, we'll discuss the implementation details of each framework component, including training specifics and network architectures.

\subsection{Animatable 3D GAN}

\begin{figure}[t]
  \centering
  \includegraphics[width=0.45\textwidth]{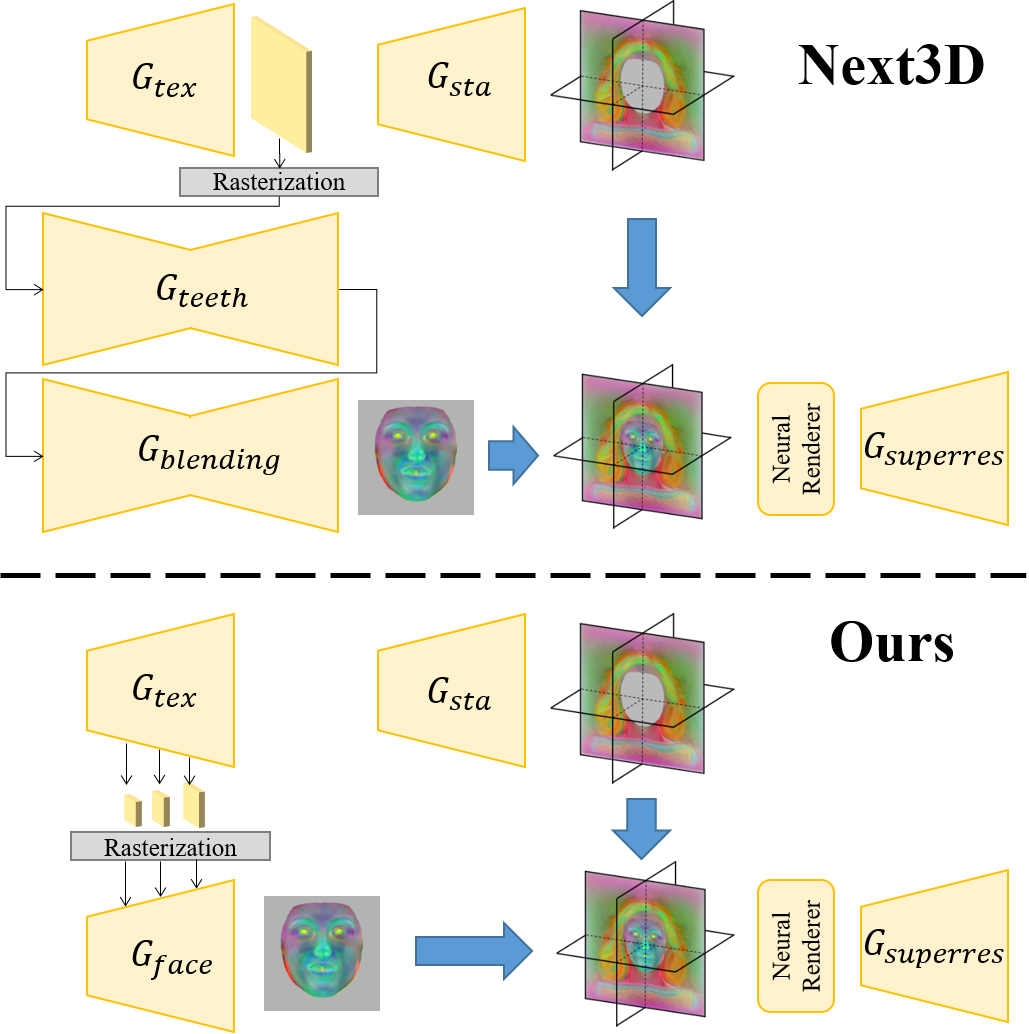}
  \caption{
  Comparison of the network architecture of Next3D against ours. Our more memory-friendly and compact representation allows for efficient training at a neural rendering resolution of $128^2$, which significantly improves view consistency.
}
  \label{fig:3dgan}
\end{figure}

\begin{figure}[t]
  \centering
  \includegraphics[width=0.35\textwidth]{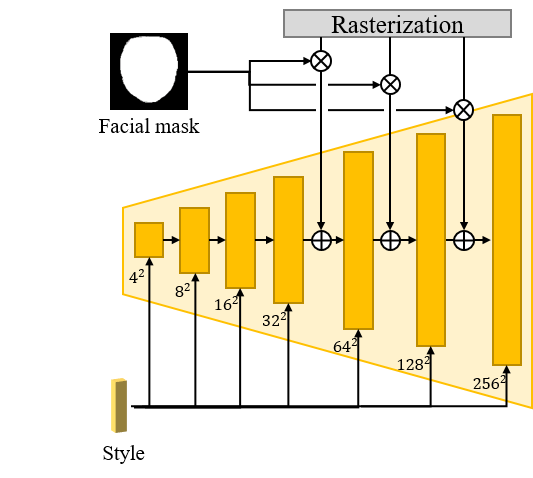}
  \caption{
  The detailed architecture of $G_{face}$.
}
  \label{fig:G_face}
\end{figure}

We implement our 3D GAN framework on top of the official PyTorch implementation of Next3D~\cite{sun2023next3d}, and we illustrate the differences in network architecture between ours and Next3D in Fig.~\ref{fig:3dgan}.
Next3D transforms neural texture into orthogonal views and forms tri-plane features. To compensate for the FLAME model's missing inner mouth area, Next3D utilizes a UNet $G_{teeth}$ to supplement and reintegrate this texture, using another UNet $G_{blending}$ for post-blending. These operations are redundant and lead to heavy parameter size, so we alternately propose a more compact architecture, maintaining the insight of completing inner mouth feature conditioning on global information, but removing heavy networks for specific mouth synthesis. Specifically, we propose a face synthesis module $G_{face}$ which is conditioned on the multi-scale rasterized neural textures. We illustrate its network architecture in Fig.~\ref{fig:G_face}. The multi-scale neural textures are rasterized from texture space into tri-plane space and then integrated with $G_{face}$'s corresponding feature maps through alpha blending. Conditioned on such global information, $G_{face}$ is precisely controlled by the deformable mesh model, and gradually inpaint inner mouth features in the forward pass layer by layer. Additionally, our more memory-efficient and compact model design allows for neural rendering resolution of $128^2$, significantly enhancing view-consistency.
Specifically, we rasterize neural texture feature maps at the resolutions of ($32, 64, 128$) for multi-scale conditions on $G_{face}$.

\subsubsection{Training Details}

We adopt many hyperparameters and training strategies of EG3D~\cite{eg3d2022} and Next3D~\cite{sun2023next3d} including blurred real images at the beginning, pose conditioned generator, density regularization, learning rates, and two-stage training. As for the deformation-aware discriminator, we replace the input synthetic rendering image with the landmarks counter map to force the generator to focus on the expression changes in the mouth and eyes region.
During the training process, we follow Next3D and adopt the same loss terms. Specifically, we train the network on FFHQ~\cite{Karras2020stylegan2}, using the non-saturating GAN loss with R1 regularization and the density regularization proposed in EG3D. 
Based on the pretrained model of EG3D, we train our model at a neural rendering resolution of 64 on four 3090 GPUs with a batch size of 16 for roughly 2 days and gradually step the resolution up to 128 for additional 2 days.

\subsection{Training Encoders}
To promote inversion performance and minimize GPU memory usage during training, we adopt a three-stage training schedule, which separately trains the latent encoder $E_{latent}$, image encoders $E_{tex}$ and $E_{tri}$ in a one-shot setting, and temporal aggregation encoders $E_{tex\_rec}$ and $E_{tri\_rec}$ in a multi-shot setting. 

\subsubsection{Training $E_{latent}$}
\label{sec:training_e4e}
For training the latent encoder $E_{latent}$, we employ pixel-wise $L_{1}$ loss, perceptual loss $L_{lpips}$~\cite{zhang2018perceptual}, and identity similarity loss $L_{ID}$~\cite{deng2018arcface}. Besides, we follow~\cite{tov2021e4e} and set a latent discriminator $D_{W}$ to discriminate between real samples from the $W$ space and the latent codes predicted by the encoder. The encoder is trained in an adversarial scheme and the total loss can be written as :
\begin{equation}
\label{eq:loss_e4e}
    L_{stage1} = L_{1} + \lambda_{lpips}L_{lpips} + \lambda_{id}L_{ID} + L_{adv}^{D_{W}} + L_{adv}^{E_{latent}}
\end{equation}
where $\lambda_{lpips}=0.5, \lambda_{id}=0.25$. The learning rates for the encoder and latent discriminator are both 1e-4. Based on the FFHQ dataset, the training is conducted on four 3090 GPUs with a batch size of 16 for 2 days.

\subsubsection{Training $E_{tri}$ and $E_{tex}$}
\label{sec:training_os}
In this stage, we fix the latent encoder $E_{latent}$ and train both UNet-style networks $E_{tri}$ and $E_{tex}$ with image datasets. Besides the FFHQ dataset, we also utilize the synthetic data generated by the pretrained GAN to prompt multi-view consistent reconstruction.

Inspired by~\cite{trevithick2023}, leveraging synthetic data, we supervise the training with intermediate feature maps, including neural texture maps $F_{tex}$, tri-plane feature maps $F_{tri}$ and neural rendering feature maps $I_{raw}$. Loss can be written as:
\begin{equation}
\label{eq:loss_os}
\begin{split}
    L_{stage2} = & L_{1} + \lambda_{lpips}L_{lpips} + \lambda_{tri}L_{tri} + \lambda_{tex}L_{tex} \\
    & + \lambda_{raw}L_{raw} + \lambda_{adv}L_{adv}
\end{split}
\end{equation}
where $\lambda_{lpips}=1.0$, $\lambda_{tri}=0.001$, $\lambda_{tex}=0.001$, $\lambda_{raw}=1.0$, $\lambda_{adv}=0.1$. $L_{tri}$, $L_{tex}$, $L_{raw}$ are the L1 losses that intuitively compare those quantities synthesized by our generative model and those created by our encoders. $L_{adv}$ is the adversarial loss using a dual discriminator~\cite{eg3d2022}, which is trained to differentiate between synthetic images sampled from GAN prior and the corresponding image reconstructed by the encoders. We remove the pose condition for the dual discriminator.
On four 3090 GPUs with a batch size of 16, we first train both image encoders without the generative adversarial objective for about 4 days, and then add adversarial loss to finetune encoders for an additional day. We use a learning rate of 1e-4 for encoders and 1e-3 for the discriminator.

\subsubsection{Training $E_{tri\_rec}$ and $E_{tex\_rec}$}
Building upon pretrained $E_{tri}$ and $E_{tex}$, $E_{tri\_rec}$ and $E_{tex\_rec}$ maintain fixed encoding backbone and only finetune their recurrent decoders. To learn to aggregate temporal information from monocular videos, we leverage  7500 real-world video clips from the VFHQ video dataset~\cite{xie2022vfhq}, each comprising 100 frames. However, since the image quality in the video dataset typically lags behind that of the GAN prior from the FFHQ dataset, training only with the VFHQ data causes blurry and degraded renderings. Hence during training, we simultaneously use the synthetic video data, which has random identity sampled from the GAN latent space, with poses and expressions sampled from the VFHQ video dataset. This additional synthetic data supplementation improves the quality of the generated images. 

Due to GPU memory constraints, during training, the temporal aggregation networks $E_{tri\_rec}$ and $E_{tex\_rec}$ process a maximum of 4 frames at a time in the forward pass. This limitation can lead to overfitting on short sequences, hindering the efficient fusion of observations from longer sequences. Leveraging the RNN framework's support for sequential inputs, we can initialize $h_0$ (Eq.1 in the main paper) to a non-zero state by cycling through the inputs during training. 
In practice, we input sequences of up to 32 frames~\footnote{Longer input is allowed. But for 100-frame video clips in our training data, it's dense enough to sample 32 frames}, randomly selecting 4 frames for rendering and loss calculation in each iteration. This strategy simulates the supervision over longer sequences to some extent, enhancing the ability of the update gates in GRU blocks to capture long-term dependencies and allowing the temporal aggregation networks to retain long-term temporal information.




We adopt the same loss terms and training strategy as in Sec~\ref{sec:training_os}. On eight 3090 GPUs with a batch size of 8, we train both encoders without the adversarial objective for about 3 days, and then finetune with adversarial loss for additional 24 hours. 

\subsection{Network Architecture}

\paragraph{Latent Encoder} 
To implement the latent encoder $E_{latent}$, we adopt the design of the e4e encoder~\cite{tov2021e4e} and employ IR-SE-50~\cite{deng2018arcface} pretrained model for the backbone network.

\paragraph{One-shot Image Encoders}
Adopting conventional U-Net architecture, $E_{tri}$ and $E_{tex}$ are designed based on TriplaneNet~\cite{bhattarai2023triplanenet}. As both networks produce multi-resolution feature maps, we incorporate an extra MLP block at each layer to transform the output channels.

\paragraph{Temporal Aggregation Networks} 
$E_{tex\_rec}$ and $E_{tri\_tex}$ are modified based on $E_{tex}$ and $E_{tri}$. Specifically, we fix the down-sampling encoding part of the networks and replace the decoding parts with recurrent decoders by injecting a ConvGRU block at each layer.


\begin{figure}[t]
  \centering
  \includegraphics[width=0.4\textwidth]{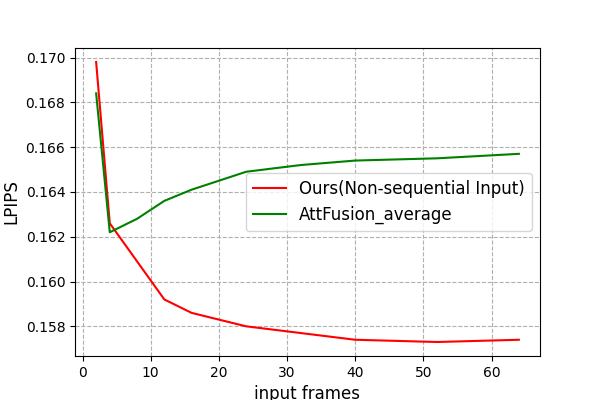}
  \caption{
  LPIPS over the number of input frames on VFHQ-test. We demonstrate the effectiveness of our method to handle unordered input. Besides, we conduct a comparison with an Attention-model baseline to validate our choice of recurrent network.
}
  \label{fig:add_att}
\end{figure}

\section{More Experiments}
\subsection{Reason for the choice of Recurrent Network}

We choose RNN for two reasons. Firstly, for few-shot tasks, the length of the input image sequence usually does not exceed 50, and GRU can handle such short sequences without suffering from forgetting issues. Secondly, despite Attention's widespread use for unordered sequences, its high computational cost limits supported sequence lengths during training. In contrast, GRU's computation grows linearly with sequence length, allowing the utilization of more frames for learning long-term temporal aggregation during training.

To evaluate our recurrent mechanism, we establish a fixed-window baseline termed ``AttFusion'', which replaces each GRU block with a transformer block to merge temporal features. As we can only train it with 3 frames in our practice, during inference ``AttFusion'' averages the inverted canonical features when handling longer sequences.

To demonstrate our ability to handle non-sequential input, we randomly sample from a video to form input sequences. 
It can be observed that the two curves representing our method in Fig.~\ref{fig:add_att} and Fig.~\ref{fig:abl_multiT} almost overlap, indicating that our approach is comparably effective in handling both ordered and unordered data.

As illustrated in Fig.~\ref{fig:add_att}, Our ConvGRU-based implementation and Attention models have similar performance for short sequences. However, for longer sequences, the performance of the Attention baseline suffers from the blur caused by the averaging strategy, and our recurrent framework's performance improves incrementally with additional frames. Hence, we think ConvGRU is more suitable for our flexible few-shot task. 

\subsection{Ablation on the network design of Animatable 3D GAN}

As mentioned in Sec.~3.2 and Sec.~A.1, we replace the two UNets in Next3D with a single generator $G_{face}$. To evaluate the impact of the replacement, we use the FFHQ dataset to train Next3D network with the same facial parametric model (FaceVerse), and illustrate the quantitative comparison in Tab.~\ref{table:cmp_gface}.

Despite a slight increase in FID score due to fewer parameters, the improvement in AKD score demonstrates that our generator has more precise expression-driving performance. This is attributed to $G_{face}$’s multi-scale conditioning, which enables the conditional parametric model to directly control generated facial features. In contrast, the two cascaded post-processing UNets diminish control sensitivity.

\begin{table}[t]
    \centering
    \begin{tabular}{c|cccc}
    \hline
        ~ & FID$\downarrow$ & AKD$\downarrow$ & CSIM$\uparrow$ & Params \\ \hline
        Ours & 4.1  & 0.057 & 0.853 & 84.21M \\
        Next3D (FV)  & 3.9  & 0.095 & 0.852 & 164.81M \\ \hline
    \end{tabular}
\caption{Quantitative comparison with Next3D (FaceVerse) on image synthesis quality, animation accuracy, identity preservation and memory consumption.}
\label{table:cmp_gface}
\end{table}







\begin{figure*}[ht]
  \centering
  \includegraphics[width=\textwidth]{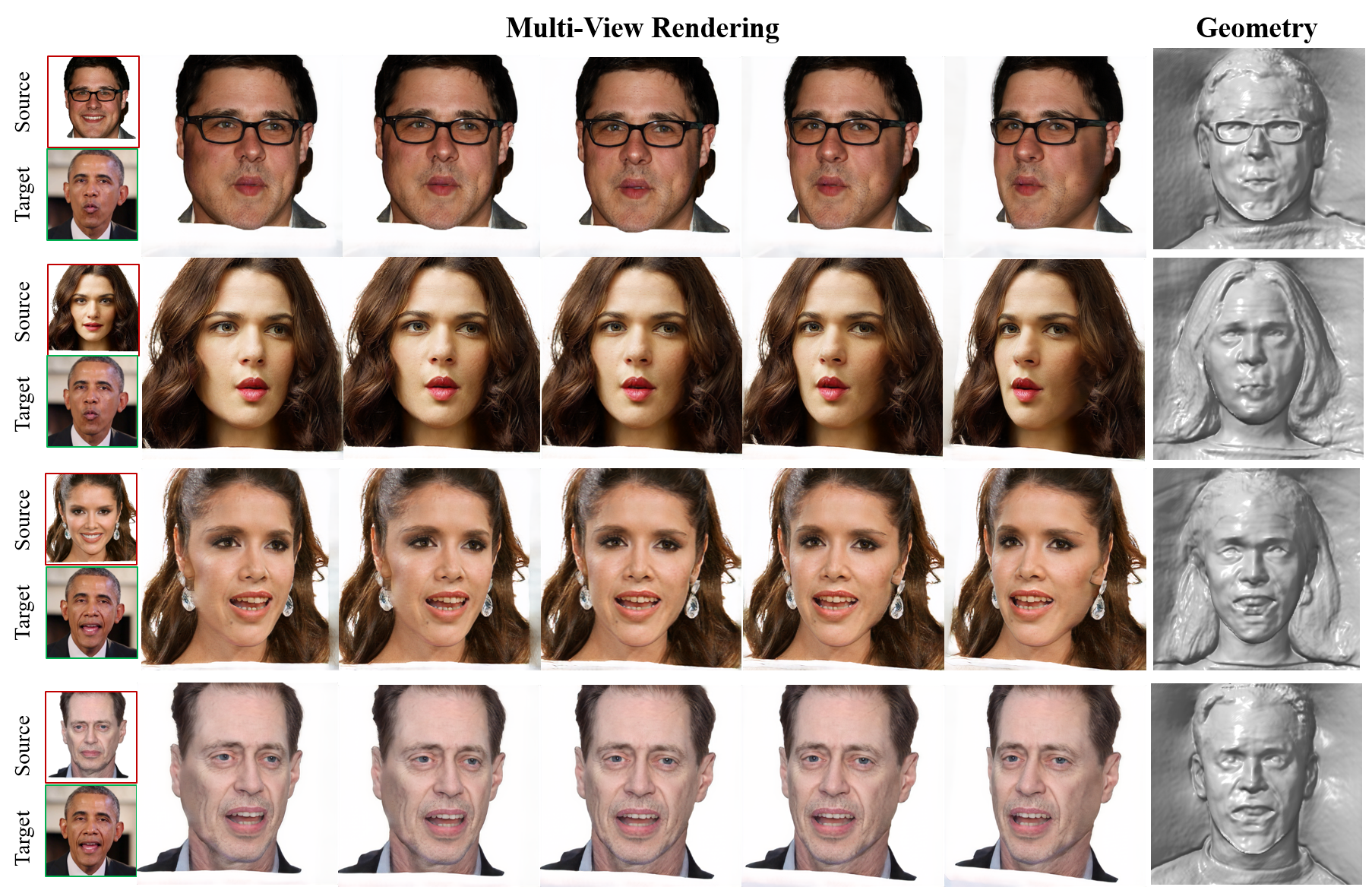}
  \caption{Cross-identity reenactment results of one-shot avatar reconstruction. We illustrate the multi-view rendering results and geometry.
}
  \label{fig:supp_geo}
\end{figure*}

\end{document}